\documentclass[11pt,a4paper]{article}
\usepackage[hyperref]{acl2021}
\usepackage{times}
\usepackage{latexsym}

\usepackage{microtype}

\usepackage{hyperref}
\usepackage{url}
\usepackage{multirow,booktabs, hhline}
\usepackage[ruled,noend]{algorithm2e}
\usepackage{amsmath, bm}
\SetKwInput{KwInput}{Input}
\SetKwInput{KwOutput}{Output}
\usepackage{url}

\usepackage{wrapfig}
\usepackage{lipsum}
\usepackage{caption}
\usepackage{float}
\usepackage{wrapfig}

\usepackage{color}
\usepackage{graphicx}
\usepackage{amsmath}
\usepackage{amssymb}
\usepackage{caption}
\usepackage{algorithmicx}
\usepackage{algpseudocode}
\usepackage{multicol}
\usepackage{multirow}
\usepackage{booktabs}
\usepackage{adjustbox}
\usepackage{longtable}
\usepackage{listings}
\usepackage{url}
\usepackage{bm}
\usepackage{xspace}
\usepackage{makecell}
\usepackage{lipsum}
\usepackage{comment}
\usepackage{subcaption}

\aclfinalcopy 
\def\aclpaperid{2129} 

\newcommand{\model}{CLEVE\xspace}

\title{\model: Contrastive Pre-training for Event Extraction}

\author{Ziqi Wang$^{1}$\thanks{\quad indicates equal contribution}, Xiaozhi Wang$^{1*}$, Xu Han$^{1}$, \textbf{Yankai Lin}$^{3}$, \textbf{Lei Hou}$^{1,2}$\thanks{\quad Correspondence to L.Hou (houlei@tsinghua.edu.cn)}\hspace{0.5em}, \\ \textbf{Zhiyuan Liu}$^{1,2}$, \textbf{Peng Li}$^{3}$, \textbf{Juanzi Li}$^{1,2}$, \textbf{Jie Zhou}$^{3}$\\
$^{1}$Department of Computer Science and Technology, BNRist;\\
$^{2}$KIRC, Institute for Artificial Intelligence, \\ Tsinghua University, Beijing, 100084, China\\
$^{3}$Pattern Recognition Center, WeChat AI, Tencent Inc, China\\
\texttt{\{ziqi-wan16, wangxz20, hanxu17\}@mails.tsinghua.edu.cn}\\ 
 }

\date{}

\begin{document}
\maketitle

\begin{abstract}

Event extraction (EE) has considerably benefited from pre-trained language models (PLMs) by fine-tuning. However, existing pre-training methods have not involved modeling event characteristics, resulting in the developed EE models cannot take full advantage of large-scale unsupervised data. To this end, we propose CLEVE, a contrastive pre-training framework for EE to better learn event knowledge from large unsupervised data and their semantic structures (e.g. AMR) obtained with automatic parsers. CLEVE contains a text encoder to learn event semantics and a graph encoder to learn event structures respectively. Specifically, the text encoder learns event semantic representations by self-supervised contrastive learning to represent the words of the same events closer than those unrelated words; the graph encoder learns event structure representations by graph contrastive pre-training on parsed event-related semantic structures. The two complementary representations then work together to improve both the conventional supervised EE and the unsupervised ``liberal'' EE, which requires jointly extracting events and discovering event schemata without any annotated data. Experiments on ACE 2005 and MAVEN datasets show that CLEVE achieves significant improvements, especially in the challenging unsupervised setting. The source code and pre-trained checkpoints can be obtained from \url{https://github.com/THU-KEG/CLEVE}.

\end{abstract}

\section{Introduction}



\begin{figure}[t]
\centering
\includegraphics[width = 0.48 \textwidth]{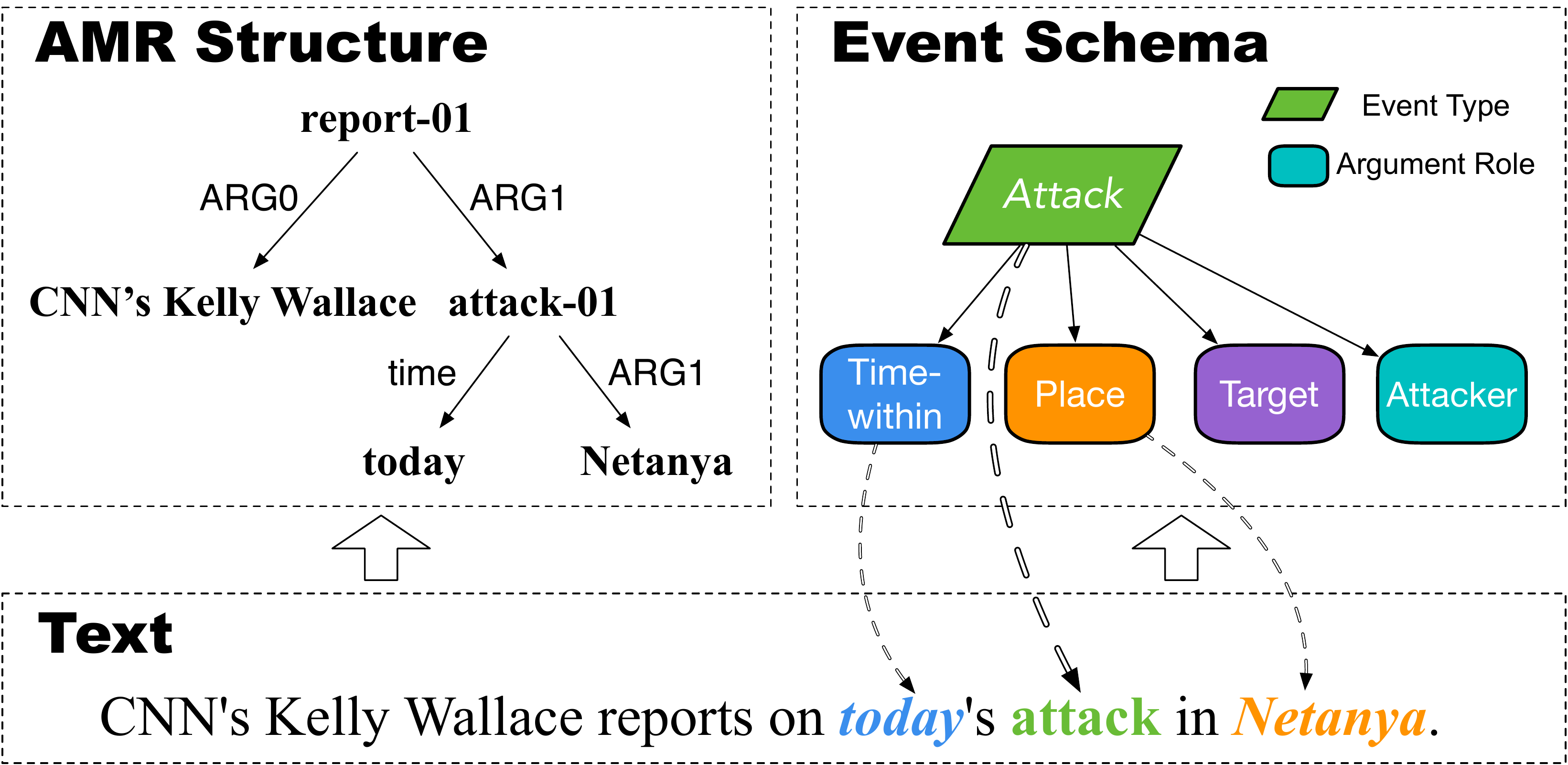}
\caption{An example sampled from the ACE 2005 dataset with its event annotation and AMR structure.}
\label{fig:EEexample}
\end{figure}

Event extraction (EE) is a long-standing crucial information extraction task, which aims at extracting event structures from unstructured text. As illustrated in Figure~\ref{fig:EEexample}, it contains event detection task to identify event triggers (the word ``attack'') and classify event types (\texttt{Attack}), as well as event argument extraction task to identify entities serving as event arguments (``today'' and ``Netanya'') and classify their argument roles (\texttt{Time-within} and \texttt{Place})~\cite{ahn2006stages}. By explicitly capturing the event structure in the text, EE can benefit various downstream tasks such as information retrieval~\cite{glavavs2014event} and knowledge base population~\cite{ji-grishman-2011-knowledge}. 

Existing EE methods mainly follow the supervised-learning paradigm to train advanced neural networks~\cite{chen2015event,nguyen2016joint,nguyen2018graph} with human-annotated datasets and pre-defined event schemata. These methods work well in lots of public benchmarks such as ACE 2005~\cite{walker2006ace} and TAC KBP~\cite{ellis2016overview}, yet they still suffer from data scarcity and limited generalizability. Since annotating event data and defining event schemata are especially expensive and labor-intensive, existing EE datasets typically only contain thousands of instances and cover limited event types. Thus they are inadequate to train large neural models~\cite{wang-etal-2020-maven} and develop methods that can generalize to continually-emerging new event types~\cite{huang-ji-2020-semi}.

Inspired by the success of recent pre-trained language models (PLMs) for NLP tasks, some pioneering work~\cite{wang-etal-2019-adversarial-training,wadden-etal-2019-entity} attempts to fine-tune general PLMs (e.g, BERT~\cite{devlin-etal-2019-bert}) for EE. Benefiting from the strong general language understanding ability learnt from large-scale unsupervised data, these PLM-based methods have achieved state-of-the-art performance in various public benchmarks.

Although leveraging unsupervised data with pre-training has gradually become a consensus for EE and NLP community, there still lacks a pre-training method orienting event modeling to take full advantage of rich event knowledge lying in large-scale unsupervised data. The key challenge here is to find reasonable self-supervised signals~\cite{chen-etal-2017-automatically,wang-etal-2019-adversarial-training} for the diverse semantics and complex structures of events. Fortunately, previous work~\cite{aguilar-etal-2014-comparison,huang-etal-2016-liberal} has suggested that sentence semantic structures, such as abstract meaning representation (AMR)~\cite{banarescu-etal-2013-AMR}, contain broad and diverse semantic and structure information relating to events. As shown in Figure~\ref{fig:EEexample}, the parsed AMR structure covers not only the annotated event (\texttt{Attack}) but also the event that is not defined in the ACE 2005 schema (\texttt{Report}). 

Considering the fact that the AMR structures of large-scale unsupervised data can be easily obtained with automatic parsers~\cite{wang-etal-2015-boosting}, we propose CLEVE, an event-oriented contrastive pre-training framework utilizing AMR structures to build self-supervision signals. CLEVE consists of two components, including a text encoder to learn event semantics and a graph encoder to learn event structure information.
Specifically, to learn effective event semantic representations, we employ a PLM as the text encoder and encourage the representations of the word pairs connected by the \texttt{ARG}, \texttt{time}, \texttt{location} edges in AMR structures to be closer in the semantic space than other unrelated words, since these pairs usually refer to the trigger-argument pairs of the same events (as shown in Figure~\ref{fig:EEexample})~\cite{huang-etal-2016-liberal}. This is done by contrastive learning with the connected word pairs as positive samples and unrelated words as negative samples. Moreover, considering event structures are also helpful in extracting events~\cite{lai-etal-2020-event} and generalizing to new event schemata~\cite{huang-etal-2018-zero}, we need to learn transferable event structure representations. Hence we further introduce a graph neural network (GNN) as the graph encoder to encode AMR structures as structure representations. The graph encoder is contrastively pre-trained on the parsed AMR structures of large unsupervised corpora with AMR subgraph discrimination as the objective.

By fine-tuning the two pre-trained models on downstream EE datasets and jointly using the two representations, CLEVE can benefit the conventional supervised EE suffering from data scarcity. Meanwhile, the pre-trained representations can also directly help extract events and discover new event schemata without any known event schema or annotated instances, leading to better generalizability. This is a challenging unsupervised setting named ``liberal event extraction''~\cite{huang-etal-2016-liberal}. Experiments on the widely-used ACE 2005 and the large MAVEN datasets indicate that CLEVE can achieve significant improvements in both settings.

\section{Related Work}
\paragraph{Event Extraction.}
Most of the existing EE works follow the supervised learning paradigm. Traditional EE methods~\cite{ji2008refining,gupta-ji:2009:Short,li2013joint} rely on manually-crafted features to extract events. In recent years, the neural models become mainstream, which automatically learn effective features with neural networks, including convolutional neural networks~\cite{nguyen-grishman-2015-event,chen2015event}, recurrent neural networks~\cite{nguyen2016joint}, graph convolutional networks~\cite{nguyen2018graph,lai-etal-2020-event}. With the recent successes of BERT~\cite{devlin-etal-2019-bert}, PLMs have also been used for EE~\cite{wang-etal-2019-adversarial-training,wang-etal-2019-hmeae,yang-etal-2019-exploring-pre,wadden-etal-2019-entity,tong-etal-2020-improving}. Although achieving remarkable performance in benchmarks such as ACE 2005~\cite{walker2006ace} and similar datasets~\cite{ellis2015overview,ellis2016overview,getman2017overview,wang-etal-2020-maven}, these PLM-based works solely focus on better fine-tuning rather than pre-training for EE. In this paper, we study pre-training to better utilize rich event knowledge in large-scale unsupervised data.

\begin{figure*}[t]
\centering
\small
\includegraphics[width = 0.98\textwidth]{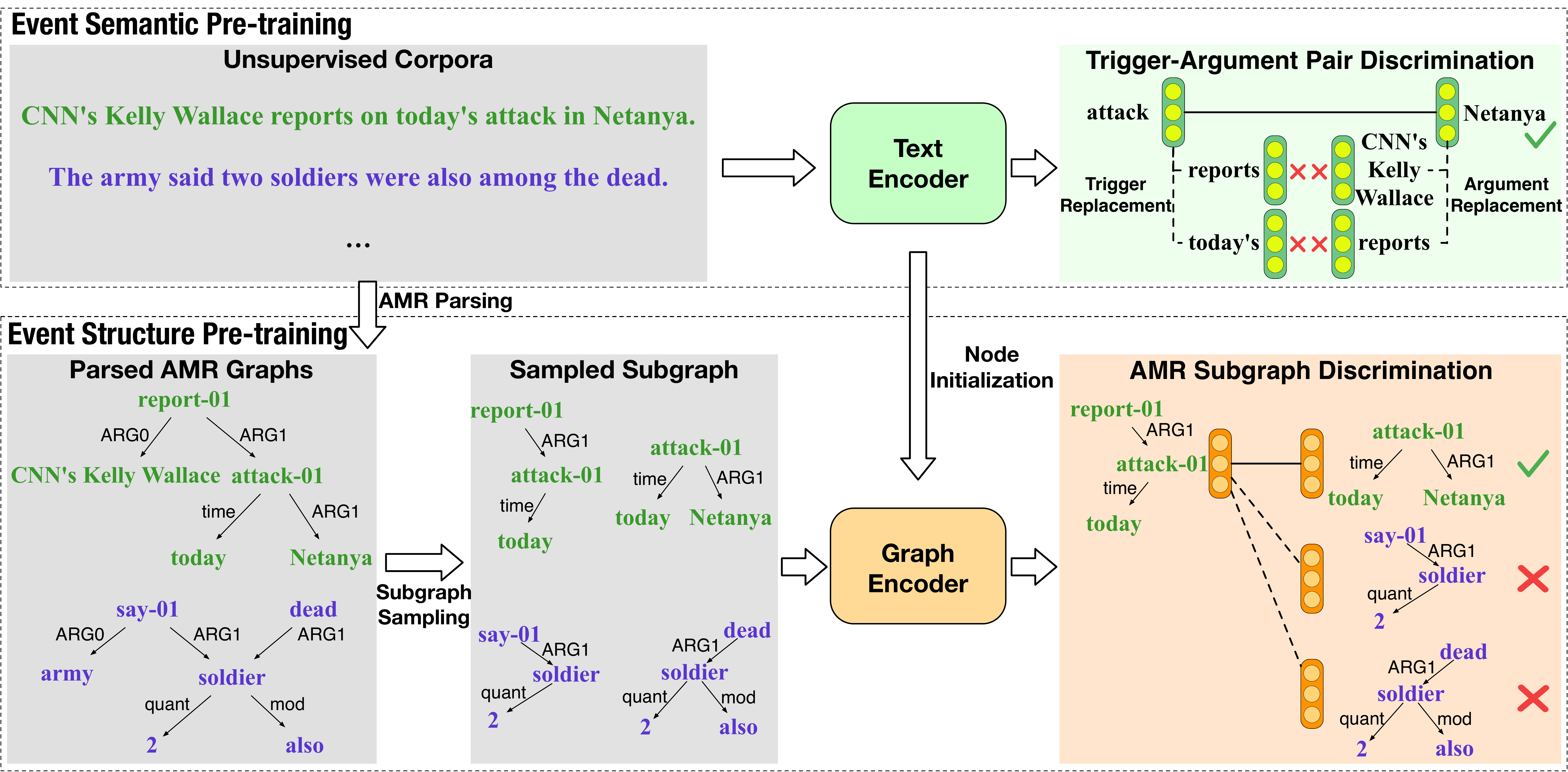}
\caption{Overall CLEVE framework. Best viewed in color.}
\label{fig:framework}
\end{figure*}

\paragraph{Event Schema Induction.}
Supervised EE models cannot generalize to continually-emerging new event types and argument roles. To this end,~\citet{chambers-jurafsky-2011-template} explore to induce event schemata from raw text by unsupervised clustering. Following works introduce more features like coreference chains~\cite{chambers-2013-event} and entities~\cite{nguyen-etal-2015-generative,sha-etal-2016-joint}. Recently, \citet{huang-ji-2020-semi} move to the semi-supervised setting allowing to use annotated data of known types. Following~\citet{huang-etal-2016-liberal}, we evaluate the generalizability of CLEVE in the most challenging unsupervised ``liberal'' setting, which requires to induce event schemata and extract event instances only from raw text at the same time.

\paragraph{Contrastive Learning.}
Contrastive learning was initiated by~\citet{Hadsell2006CL} following an intuitive motivation to learn similar representations for ``neighboors'' and distinct representations for ``non-neighbors'', and is further widely used for self-supervised representation learning in various domains, such as computer vision~\cite{wu2018unsupervised,oord2018representation,hjelm2019DIM,chen2020SIMCLR,he2020momentum} and graph~\cite{qiu2020GCC,you2020graph,Zhu:2020vf}.
In the context of NLP, many established representation learning works can be viewed as contrastive learning methods, such as Word2Vec~\cite{mikolov2013word2vec}, BERT~\cite{devlin-etal-2019-bert,kong2020mutual} and ELECTRA~\cite{clark2020electra}. Similar to this work, contrastive learning is also widely-used to help specific tasks, including question answering~\cite{yeh-chen-2019-qainfomax}, discourse modeling~\cite{iter-etal-2020-pretraining}, natural language inference~\cite{cui-etal-2020-unsupervised} and relation extraction~\cite{peng-etal-2020-learning}.
\section{Methodology}

The overall CLEVE framework is illustrated in Figure~\ref{fig:framework}. As shown in the illustration, our contrastive pre-training framework CLEVE consists of two components: event semantic pre-training and event structure pre-training, of which details are introduced in Section~\ref{sec:semantic} and Section~\ref{sec:structure}, respectively. At the beginning of this section, we first introduce the required preprocessing in Section~\ref{sec:preliminary}, including the AMR parsing and how we modify the parsed AMR structures for our pre-training.

\subsection{Preprocessing}
\label{sec:preliminary}
CLEVE relies on AMR structures~\cite{banarescu-etal-2013-AMR} to build broad and diverse self-supervision signals for learning event knowledge from large-scale unsupervised corpora. To do this, we use automatic AMR parsers~\cite{wang-etal-2015-boosting,xu-etal-2020-improving} to parse the sentences in unsupervised corpora into AMR structures. Each AMR structure is a directed acyclic graph with concepts as nodes and semantic relations as edges. Moreover, each node typically only corresponds to at most one word, and a multi-word entity will be represented as a list of nodes connected with \texttt{name} and \texttt{op} (conjunction operator) edges. Considering pre-training entity representations will naturally benefits event argument extraction, we merge these lists into single nodes representing multi-word entities (like the ``CNN's Kelly Wallace'' in Figure~\ref{fig:EEexample}) during both event semantic and structure pre-training. Formally, given a sentence $s$ in unsupervised corpora, we obtain its AMR graph $g_{s}=\left(\mathrm{V}_{s}, \mathrm{E}_{s}\right)$ after AMR parsing, where $\mathrm{V}_{s}$ is the node set after word merging and $\mathrm{E}_{s}$ denotes the edge set. $\mathrm{E}_{s}=\{\left(u,v,r\right)|\left(u,v\right) \in \mathrm{V}_{s} \times \mathrm{V}_{s}, r \in \mathcal{R}\}$, where $\mathcal{R}$ is the set of defined semantic relation types.

\subsection{Event Semantic Pre-training}
\label{sec:semantic}
To model diverse event semantics in large unsupervised corpora and learn contextualized event semantic representations, we adopt a PLM as the text encoder and train it with the objective to discriminate various trigger-argument pairs.
\subsubsection*{Text Encoder}
\label{sec:text_encoder}
Like most PLMs, we adopt a multi-layer Transformer~\cite{vaswani2017attention} as the text encoder since its strong representation capacity. Given a sentence $s = \{w_1, w_2,\ldots, w_n\}$ containing $n$ tokens, we feed it into the multi-layer Transformer and use the last layer's hidden vectors as token representations. 
Moreover, a node $v \in \mathrm{V}_{s}$ may correspond to a multi-token text span in $s$ and we need a unified representation for the node in pre-training. As suggested by~\citet{baldini-soares-etal-2019-matching}, we insert two special markers \texttt{[E1]} and \texttt{[/E1]} at the beginning and ending of the span, respectively. Then we use the hidden vector for \texttt{[E1]} as the span representation $\mathbf{x}_{v}$ of the node $v$. And we use different marker pairs for different nodes. 

As our event semantic pre-training focuses on modeling event semantics, we start our pre-training from a well-trained general PLM to obtain general language understanding abilities. CLEVE is agnostic to the model architecture and can use any general PLM, like BERT~\cite{devlin-etal-2019-bert} and RoBERTa~\cite{liu2019roberta}.

\subsubsection*{Trigger-Argument Pair Discrimination}

We design trigger-argument pair discrimination as our contrastive pre-training task for event semantic pre-training. The basic idea is to learn closer representations for the words in the same events than the unrelated words. We note that the words connected by \texttt{ARG}, \texttt{time} and \texttt{location} edges in AMR structures are quite similar to the trigger-argument pairs in events~\cite{huang-etal-2016-liberal,huang-etal-2018-zero}, i.e., the key words evoking events and the entities participating events. For example, in Figure~\ref{fig:EEexample}, ``Netanya'' is an argument for the ``attack'' event, while the disconnected ``CNN's Kelly Wallace'' is not. With this observation, we can use these special word pairs as positive trigger-argument samples and train the text encoder to discriminate them from negative samples, so that the encoder can learn to model event semantics without human annotation.

Let $\mathcal{R}_{p}=\{\texttt{ARG}, \texttt{time}, \texttt{location} \}$ and $\mathrm{P}_{s}=\{(u,v)|\exists (u,v,r) \in \mathrm{E}_{s}, r\in \mathcal{R}_{p}\}$ denotes the set of positive trigger-argument pairs in sentence $s$. For a specific positive pair $(t,a)\in\mathrm{P}_{s}$, as shown in Figure~\ref{fig:framework}, we construct its corresponding negative samples with trigger replacement and argument replacement. Specifically, in the trigger replacement, we construct $m_{t}$ number of negative pairs by randomly sample $m_t$ number of negative triggers $\hat{t} \in \mathrm{V}_{s}$ and combine them with the positive argument $a$. A negative trigger $\hat{t}$ must do not have a directed \texttt{ARG}, \texttt{time} or \texttt{location} edge with $a$, i.e., $\nexists (\hat{t},a,r)\in\mathrm{E}_{s},r\in\mathcal{R}_{p}$. Similarly, we construct $m_{a}$ more negative pairs by randomly sample $m_a$ number of negative arguments $\hat{a} \in \mathrm{V}_{s}$ satisfying $\nexists (t,\hat{a},r)\in\mathrm{E}_{s},r\in\mathcal{R}_{p}$. As the example in Figure~\ref{fig:framework}, (``attack'', ``reports'') is a valid negative sample for the positive sample (``attack'', ``Netanya''), but (``attack'', ``today's'') is not valid since there is a (``attack'', ``today's'', \texttt{time}) edge. 

To learn to discriminate the positive trigger-argument pair from the negative pairs and so that model event semantics, we define the training objective for a positive pair $(t,a)$ as a cross-entropy loss of classifying the positive pair correctly:
\begin{equation}
\small
\begin{aligned}
\mathcal{L}_{t,a}=&-\mathbf{x}_{t}^{\top}W\mathbf{x}_{a}\\
&+\mathrm{log}\Big(\mathrm{exp}\left(\mathbf{x}_{t}^{\top}W\mathbf{x}_{a}\right)+\sum_{i=1}^{m_{t}}\mathrm{exp}\left(\mathbf{x}_{\hat{t}_{i}}^{\top}W\mathbf{x}_{a}\right)\\&+\sum_{j=1}^{m_{a}}\mathrm{exp}\left(\mathbf{x}_{t}^{\top}W\mathbf{x}_{\hat{a}_{j}}\right) \Big),
\end{aligned}
\end{equation}
where $m_t$, $m_a$ are hyper-parameters for negative sampling, and $W$ is a trainable matrix learning the similarity metric. We adopt the cross-entropy loss here since it is more effective than other contrastive loss forms~\cite{oord2018representation,chen2020SIMCLR}. 

Then we obtain the overall training objective for event semantic pre-training by summing up the losses of all the positive pairs of all sentences $s$ in the mini batch $\mathrm{B}_s$:
\begin{equation}
    \small
    \begin{aligned}
    \mathcal{L}_{sem}(\theta)=\sum_{s \in \mathrm{B}_{s}}\sum_{(t,a)\in\mathrm{P}_{s}} \mathcal{L}_{t,a},
    \end{aligned}
\end{equation}
where $\theta$ denotes the trainable parameters, including the text encoder and $W$.

\subsection{Event Structure Pre-training}
\label{sec:structure}
Previous work has shown that event-related structures are helpful in extracting new events~\cite{lai-etal-2020-event} as well as discovering and generalizing to new event schemata~\cite{huang-etal-2016-liberal,huang-etal-2018-zero,huang-ji-2020-semi}. Hence we conduct event structure pre-training on a GNN as graph encoder to learn transferable event-related structure representations with recent advances in graph contrastive pre-training~\cite{qiu2020GCC,you2020graph,Zhu:2020vf}. Specifically, we pre-train the graph encoder with AMR subgraph discrimination task. 
\subsubsection*{Graph Encoder}
\label{sec:graph_encoder}
In CLEVE, we utilize a GNN to encode the AMR (sub)graph to extract the event structure information of the text. Given a graph $g$, the graph encoder represents it with an graph embedding $\mathbf{g} = \mathcal{G}(g,\{\mathbf{x}_{v}\})$, where $\mathcal{G}(\cdot)$ is the graph encoder and $\{\mathbf{x}_{v}\}$ denotes the initial node representations fed into the graph encoder. CLEVE is agnostic to specific model architectures of the graph encoder. Here we use a state-of-the-art GNN model, Graph Isomorphism Network~\cite{xu2018how}, as our graph encoder for its strong representation ability.

We use the corresponding text span representations $\{\mathbf{x}_{v}\}$ produced by our pre-trained text encoder (introduced in Section~\ref{sec:text_encoder}) as the initial node representations for both pre-training and inference of the graph encoder. This node initialization also implicitly aligns the semantic spaces of event semantic and structure representations in CLEVE, so that can make them cooperate better.

\subsubsection*{AMR Subgraph Discrimination}
To learn transferable event structure representations, we design the AMR subgraph discrimination task for event structure pre-training. The basic idea is to learn similar representations for the subgraphs sampled from the same AMR graph by discriminating them from subgraphs sampled from other AMR graphs~\cite{qiu2020GCC}.

Given a batch of $m$ AMR graphs $\{g_1,g_2,\ldots,g_m\}$, each graph corresponds to a sentence in unsupervised corpora. For the $i$-th graph $g_i$, we randomly sample two subgraphs from it to get a positive pair $a_{2i-1}$ and $a_{2i}$. And all the subgraphs sampled from the other AMR graphs in the mini-batch serve as negative samples. Like in Figure~\ref{fig:framework}, the two green (w/ ``attack'') subgraphs are a positive pair while the other two subgraphs sampled from the purple (w/ ``solider'') graph are negative samples. Here we use the subgraph sampling strategy introduced by~\citet{qiu2020GCC}, whose details are shown in Appendix~\ref{app:subgraph_sampling}.

Similar to event semantic pre-training, we adopt the graph encoder to represent the samples $\mathbf{a}_{i}=\mathcal{G}\left(a_i,{\mathbf{x}_{v}}\right)$and define the training objective as:
\begin{equation}
\small
\begin{aligned}
\mathcal{L}_{str}(\theta)=-\sum_{i=1}^{m}\mathrm{log} \frac{\mathrm{exp}\left(\mathbf{a}_{2i-1}^{\top}\mathbf{a}_{2i}\right)}{\sum_{j=1}^{2m}\mathbf{1}_{[j\neq 2i-1]}~\mathrm{exp}\left(\mathbf{a}_{2i-1}^{\top}\mathbf{a}_{j}\right)},
\end{aligned}
\end{equation}
where $\mathbf{1}_{[j\neq 2i-1]} \in \{0,1\}$ is an indicator function evaluating to $1$ iff $j\neq 2i-1$ and $\theta$ is the trainable parameters of graph encoder.

\section{Experiment}
We evaluate our methods in both the supervised setting and unsupervised ``liberal'' setting of EE. 

\subsection{Pre-training Setup}
Before the detailed experiments, we introduce the pre-training setup of CLEVE in implementation. We adopt the New York Times Corpus (NYT)\footnote{\url{https://catalog.ldc.upenn.edu/LDC2008T19}}~\cite{sandhaus2008new} as the unsupervised pre-training corpora for CLEVE. It contains over $1.8$ million articles written and published by the New York Times between January 1, 1987, and June 19, 2007. We only use its raw text and obtain the AMR structures with a state-of-the-art AMR parser~\cite{xu-etal-2020-improving}. We choose NYT corpus because (1) it is large and diverse, covering a wide range of event semantics, and (2) its text domain is similar to our principal evaluation dataset ACE 2005, which is helpful~\cite{dontstoppretraining2020}. To prevent data leakage, we remove all the articles shown up in ACE 2005 from the NYT corpus during pre-training. Moreover, we also study the effect of different AMR parsers and pre-training corpora in Section~\ref{sec:analysis_AMR} and Section~\ref{sec:analysis_pre_domain}, respectively.

For the text encoder, we use the same model architecture as RoBERTa~\cite{liu2019roberta}, which is with $24$ layers, $1024$ hidden dimensions and $16$ attention heads, and we start our event semantic pre-training from the released checkpoint\footnote{\url{https://github.com/pytorch/fairseq}}. For the graph encoder, we adopt a graph isomorphism network~\cite{xu2018how} with $5$ layers and $64$ hidden dimensions, and pre-train it from scratch. For the detailed hyperparameters for pre-training and fine-tuning, please refer to Appendix~\ref{app:hyperparameter}.

\subsection{Adaptation of CLEVE}

As our work focuses on pre-training rather than fine-tuning for EE, we use straightforward and common techniques to adapt pre-trained CLEVE to downstream EE tasks. In the supervised setting, we adopt dynamic multi-pooling mechanism~\cite{chen2015event,wang-etal-2019-adversarial-training,wang-etal-2019-hmeae} for the text encoder and encode the corresponding local subgraphs with the graph encoder. Then we concatenate the two representations as features and fine-tune CLEVE on supervised datasets. In the unsupervised ``liberal'' setting, we follow the overall pipeline of~\citet{huang-etal-2016-liberal} and directly use the representations produced by pre-trained CLEVE as the required trigger/argument semantic representations and event structure representations. For the details, please refer to Appendix~\ref{app:adaption}.

\begin{table}[!t]
  \centering
   \small
   \setlength{\tabcolsep}{2.5pt}
   {
    \begin{tabular}{l|ccc|ccc}
    \toprule
          & \multicolumn{3}{c|}{\textbf{ED}} & \multicolumn{3}{c}{\textbf{EAE}} \\
    \midrule
    \textbf{Metric} & \textbf{P} & \textbf{R} & \textbf{F1}    & \textbf{P} & \textbf{R} & \textbf{F1} \\
    \midrule
    JointBeam & $73.7$ & $62.3$ & $67.5$ & $64.7$ & $44.4$ & $52.7$ \\
    DMCNN & $75.6$ & $63.6$ & $69.1$ & $62.2$ & $46.9$ & $53.5$ \\
    dbRNN & $74.1$ & $69.8$ & $71.9$ & $66.2$ & $52.8$ & $58.7$ \\
    GatedGCN & $\bm{78.8}$ & $76.3$ & $77.6$ & $-$ & $-$ & $-$ \\
    SemSynGTN & $-$ & $-$ & $-$ & $\bm{69.3}$ & $55.9$ & $61.9$ \\
    RCEE\_ER & $75.6$ & $74.2$ & $74.9$ & $63.0$ & $64.2$ & $\bm{63.6}$ \\
    RoBERTa & $75.1$ & $79.2$ & $77.1$ & $53.5$ & $66.8$ & $59.4$ \\
    \midrule
    CLEVE & $78.1$ & $\bm{81.5}$ & $\bm{79.8}$ & $55.4$ & $\bm{68.0}$ & $61.1$ \\
    \quad w/o semantic & $75.3$ & $79.7$ & $77.4$ & $53.8$ & $67.0$ & $59.7$ \\
    \quad w/o structure & $78.0$ & $81.1$ & $79.5$ & $55.1$ & $67.6$ & $60.7$\\
    \quad on ACE (golden) & $76.2$ & $79.8$ & $78.0$ & $54.2$ & $67.5$ & $60.1$ \\
    \quad on ACE (AMR) & $75.7$ & $79.5$ & $77.6$ & $53.6$ & $66.9$ & $59.5$\\
    \bottomrule
    \end{tabular}
    }
    \caption{Supervised EE performance (\%) of various models on ACE 2005.}
  \label{tab:supervised_ACE}%
\end{table}

\subsection{Supervised EE}
\label{sec:supervised}
\subsubsection*{Dataset and Evaluation}
We evaluate our models on the most widely-used ACE 2005 English subset~\cite{walker2006ace} and the newly-constructed large-scale MAVEN~\cite{wang-etal-2020-maven} dataset. ACE 2005 contains $599$ English documents, which are annotated with $8$ event types, $33$ subtypes, and $35$ argument roles. MAVEN contains $4,480$ documents and $168$ event types, which can only evaluate event detection. We split ACE 2005 following previous EE work~\cite{liao2010using,li2013joint,chen2015event} and use the official split for MAVEN. 
EE performance is evaluated with the performance of two subtasks: Event Detection (ED) and Event Argument Extraction (EAE). We report the precision (P), recall (R) and F1 scores as evaluation results, among which F1 is the most comprehensive metric. 

\paragraph{Baselines} We fine-tune our pre-trained \textbf{CLEVE} and set the original \textbf{RoBERTa} without our event semantic pre-training as an important baseline. To do ablation studies, we evaluate two variants of CLEVE on both datasets: the \textbf{w/o semantic} model adopts a vanilla RoBERTa without event semantic pre-training as the text encoder, and the \textbf{w/o structure} only uses the event semantic representations without event structure pre-training. 

On ACE 2005, we set two more variants to investigate the effectiveness of CLEVE. The \textbf{on ACE (golden)} model is pre-trained with the golden trigger-argument pairs and event structures of ACE 2005 training set instead of the AMR structures of NYT. Similarly, the \textbf{on ACE (AMR)} model is pre-trained with the parsed AMR structures of ACE 2005 training set. We also compare CLEVE with various baselines, including: (1) feature-based method, the top-performing \textbf{JointBeam}~\cite{li2013joint}; (2) vanilla neural model \textbf{DMCNN}~\cite{chen2015event}; (3) the model incorporating syntactic knowledge, \textbf{dbRNN}~\cite{lei2018jointly}; (4) state-of-the-art models on ED and EAE respectively, including \textbf{GatedGCN}~\cite{lai-etal-2020-event} and \textbf{SemSynGTN}~\cite{pouran-ben-veyseh-etal-2020-graph}; (5) a state-of-the-art EE model \textbf{RCEE\_ER}~\cite{liu-etal-2020-event}, which tackle EE with machine reading comprehension (MRC) techniques. The last four models adopt PLMs to learn representations.

On MAVEN, we compare CLEVE with the official ED baselines set by~\citet{wang-etal-2020-maven}, including \textbf{DMCNN}~\cite{chen2015event}, \textbf{BiLSTM}~\cite{hochreiter1997long}, \textbf{BiLSTM+CRF}, \textbf{MOGANED}~\cite{yan-etal-2019-event}, \textbf{DMBERT}~\cite{wang-etal-2019-adversarial-training}, \textbf{BERT+CRF}.

\begin{table}[t!]
\small
  \centering
    \begin{tabular}{l|ccc}
    \toprule
          & \multicolumn{3}{c}{\textbf{ED}} \\
    \midrule
    \textbf{Metric} & \textbf{P} & \textbf{R} & \textbf{F1} \\
    \midrule
    DMCNN   & $\bm{66.3}$ & $55.9$ & $60.6$ \\ 
    BiLSTM  & $59.8$ & $67.0$ & $62.8$ \\ 
    BiLSTM+CRF & $63.4$ & $64.8$ & $64.1$ \\
    MOGANED  & $63.4$ & $64.1$ & $63.8$ \\ 
    DMBERT & $62.7$ & $72.3$ & $67.1$ \\
    BERT+CRF  & $65.0$ & $70.9$ & $67.8$ \\
    RoBERTa & $64.3$ & $72.2$ & $68.0$ \\
    \midrule
    CLEVE & $64.9$ & $\bm{72.6}$ & $\bm{68.5}$  \\
    \quad w/o semantic & $64.5$ & $72.4$ & $68.2$ \\
    \quad w/o structure & $64.7$ & $72.5$ & $68.4$ \\
    \bottomrule
    \end{tabular}%
    \caption{Supervised EE performance (\%) of various models on MAVEN.}
  \label{tab:supervised_MAVEN}%
\end{table}

\subsubsection*{Evaluation Results}
The evaluation results are shown in Table~\ref{tab:supervised_ACE} and Table~\ref{tab:supervised_MAVEN}. We can observe that: (1) CLEVE achieves significant improvements to its basic model RoBERTa on both ACE 2005 and MAVEN. The p-values under the t-test are $4\times10^{-8}$, $2\times10^{-8}$ and $6\times10^{-4}$ for ED on ACE 2005, EAE on ACE 2005, and ED on MAVEN, respectively. It also outperforms or achieves comparable results with all the baselines, including those using dependency parsing information (dbRNN, GatedGCN, SemSynGTN and MOGANED). This demonstrates the effectiveness of our proposed contrastive pre-training method and AMR semantic structure. It is noteworthy that RCEE\_ER outperforms our method in EAE since its special advantages brought by reformulating EE as an MRC task to utilize sophisticated MRC methods and large annotated external MRC data. Considering that our method is essentially a pre-training method learning better event-oriented representations, CLEVE and RCEE\_ER can naturally work together to improve EE further. (2) The ablation studies (comparisons between CLEVE and its w/o semantic or structure representations variants) indicate that both event semantic pre-training and event structure pre-training is essential to our method. (3) From the comparisons between CLEVE and its variants on ACE (golden) and ACE (AMR), we can see that the AMR parsing inevitably brings data noise compared to golden annotations, which results in a performance drop. However, this gap can be easily made up by the benefits of introducing large unsupervised data with pre-training.

\subsection{Unsupervised ``Liberal'' EE}

\begin{table}[t!]
\small
  \centering
  \setlength{\tabcolsep}{2.5pt}
  {
    \begin{tabular}{l|ccc|ccc}
    \toprule
       & \multicolumn{3}{c|}{\textbf{ED}} & \multicolumn{3}{c}{\textbf{EAE}} \\
    \midrule
    \textbf{Metric (B-Cubed)} & \textbf{P} & \textbf{R} & \textbf{F1}    & \textbf{P} & \textbf{R} & \textbf{F1} \\
    \midrule
    LiberalEE & $55.7$ & $45.1$ & $49.8$ & $36.2$ & $26.5$ & $30.6$ \\
    \midrule
    RoBERTa & $44.3$ & $24.9$ & $31.9$ & $24.2$ & $17.3$ & $20.2$ \\

    RoBERTa+VGAE & $47.0$ & $26.8$ & $34.1$ & $25.6$ & $17.9$ & $21.1$  \\
    \midrule
    CLEVE & $\bm{62.0}$ & $\bm{47.3}$ & $\bm{53.7}$ & $\bm{41.6}$ & $\bm{30.3}$ & $\bm{35.1}$ \\
    \quad w/o semantic & $60.6$ & $46.2$ & $52.4$ & $40.9$ & $29.8$ & $34.5$ \\
    \quad w/o structure & $45.7$ & $25.6$ & $32.8$ & $25.0$ & $17.9$ & $20.9$ \\
    \quad on ACE (AMR) & $61.1$ & $46.7$ & $52.9$  & $41.5$ & $30.1$ & $34.9$\\
    \bottomrule
    \end{tabular}
    }
    \caption{Unsupervised ``liberal'' EE performance (\%) of various models on ACE 2005.}
  \label{tab:liberal_ACE}
\end{table}

\subsubsection*{Dataset and Evaluation}

In the unsupervised setting, we evaluate CLEVE on ACE 2005 and MAVEN with both objective automatic metrics and human evaluation. For the automatic evaluation, we adopt the extrinsic clustering evaluation metrics: B-Cubed Metrics~\cite{bagga-baldwin-1998-entity-based}, including B-Cubed precision, recall and F1. The B-Cubed metrics evaluate the quality of cluster results by comparing them to golden standard annotations and have been shown to be effective~\cite{amigo2009comparison}. For the human evaluation, we invite an expert to check the outputs of the models to evaluate whether the extracted events are complete and correctly clustered as well as whether all the events in text are discovered.

\paragraph{Baselines} We compare CLEVE with reproduced \textbf{LiberalEE}~\cite{huang-etal-2016-liberal}, \textbf{RoBERTa} and \textbf{RoBERTa+VGAE}. \textbf{RoBERTa} here adopts the original RoBERTa~\cite{liu2019roberta} without event semantic pre-training to produce semantic representations for trigger and argument candidates in the same way as CLEVE, and encode the whole sentences to use the sentence embeddings (embeddings of the starting token \texttt{<s>}) as the needed event structure representations. \textbf{RoBERTa+VGAE} additionally adopts an unsupervised model Variational Graph Auto-Encoder (VGAE)~\cite{kipf2016variational} to encode the AMR structures as event structure representations. RoBERTa+VGAE shares similar model architectures with CLEVE but is without our pre-training. Specially, for fair comparisons with LiberalEE, all the models in the unsupervised experiments adopt the same CAMR~\cite{wang-etal-2015-boosting} as the AMR parser, including CLEVE pre-training. Moreover, we also study CLEVE variants as in the supervised setting. The \textbf{w/o semantic} variant replaces the CLEVE text encoder with a RoBERTa without event structure pre-training. The \textbf{w/o structure} variant only uses CLEVE text encoder in a similar way as \textbf{RoBERTa}. The \textbf{on ACE (AMR)} model is pre-trained with the parsed AMR structures of ACE test set. As shown in~\citet{huang-etal-2016-liberal}, the AMR parsing is significantly superior to dependency parsing and frame semantic parsing on the unsupervised ``liberal'' event extraction task, hence we do not include baselines using other sentence structures in the experiments.

\begin{table}[t!]
\small
  \centering

    \begin{tabular}{l|ccc}
    \toprule
       & \multicolumn{3}{c}{\textbf{ED}} \\
    \midrule
    \textbf{Metric (B-Cubed)} & \textbf{P} & \textbf{R} & \textbf{F1}  \\
    \midrule
    RoBERTa & $32.1$ & $25.2$ & $28.2$ \\

    RoBERTa+VGAE & $37.7$ & $28.5$ & $32.5$ \\
    \midrule
    CLEVE & $\bm{55.6}$ & $\bm{46.4}$  & $\bm{50.6}$  \\
    \quad w/o semantic & $53.2$ & $44.8$ & $48.6$ \\
    \quad w/o structure & $32.8$ & $26.1$ & $29.1$ \\
    \bottomrule
    \end{tabular}
    
    \caption{Unsupervised ``liberal'' EE performance (\%) of various models on MAVEN.}
  \label{tab:liberal_MAVEN}
\end{table}

\begin{table}[t!]
\small
  \centering
  \setlength{\tabcolsep}{2.5pt}
  {
    \begin{tabular}{l|ccc|ccc}
    \toprule
       & \multicolumn{3}{c|}{\textbf{ED}} & \multicolumn{3}{c}{\textbf{EAE}} \\
    \midrule
    \textbf{Metric (Human)} & \textbf{P} & \textbf{R} & \textbf{F1}    & \textbf{P} & \textbf{R} & \textbf{F1} \\
    \midrule
    LiberalEE & $51.2$ & $46.9$ & $49.0$ & $33.5$ & $27.2$ & $30.0$ \\
    \midrule
    CLEVE & $\bm{60.4}$ & $\bm{48.4}$ & $\bm{53.7}$ & $\bm{39.4}$ & $\bm{31.1}$ & $\bm{34.8}$ \\
    \bottomrule
    \end{tabular}
    }
    \caption{Unsupervised ``liberal'' EE human-evaluation performance (\%) on ACE 2005. }
  \label{tab:liberal_ACE_human}
\end{table}

\subsubsection*{Evaluation Results}

The automatic evaluation results are shown in Table~\ref{tab:liberal_ACE} and Table~\ref{tab:liberal_MAVEN}. As the human evaluation is laborious and expensive, we only do human evaluations for CLEVE and the most competitive baseline LiberalEE on ACE 2005, and the results are shown in Table~\ref{tab:liberal_ACE_human}. We can observe that: (1) CLEVE significantly outperforms all the baselines, which shows its superiority in both extracting event instances and discovering event schemata. (2) RoBERTa ignores the structure information. Although RoBERTa+VAGE encodes event structures with VGAE, the semantic representations of RoBERTa and the structure representations of VGAE are distinct and thus cannot work together well. Hence the two models even underperform LiberalEE, while the two representations of CLEVE can collaborate well to improve ``liberal'' EE. (3) In the ablation studies, the discarding of event structure pre-training results in a much more significant performance drop than in the supervised setting, which indicates event structures are essential to discovering new event schemata.

\section{Analysis}
\subsection{Effect of Supervised Data Size}
\begin{figure}[t!]
\small
\centering
\scalebox{0.9}{
\includegraphics[width = 0.45\textwidth]{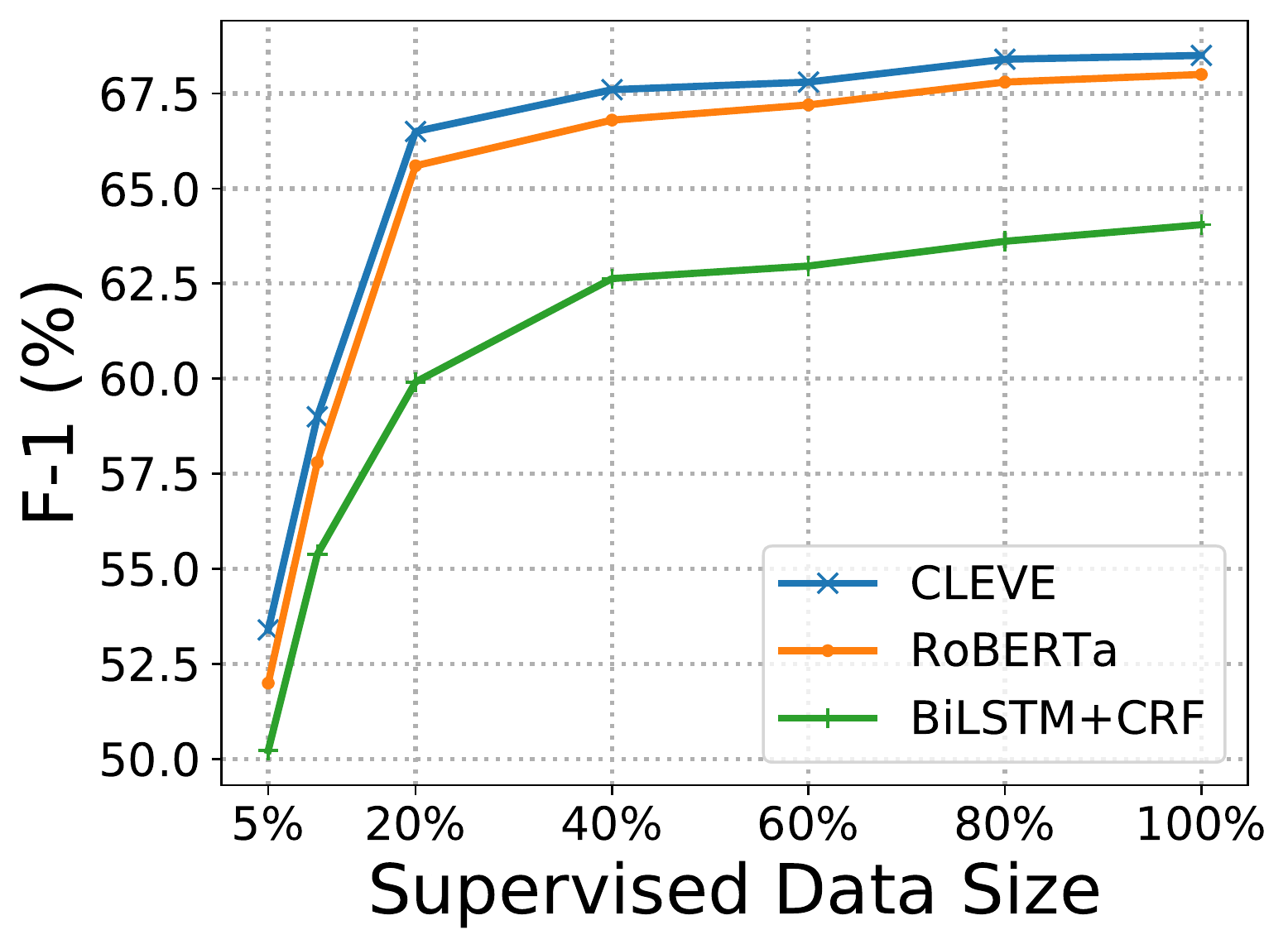}
}

\caption{Supervised ED performance (F-1) on MAVEN with different training data size.}
\label{fig:dataSize}
\end{figure}

In this section, we study how the benefits of pre-training change along with the available supervised data size. We compare the ED performance on MAVEN of CLEVE, RoBERTa and a non-pre-training model BiLSTM+CRF when trained on different proportions of randomly-sampled MAVEN training data in Figure~\ref{fig:dataSize}. We can see that the improvements of CLEVE compared to RoBERTa and the pre-training models compared to the non-pre-training model are generally larger when less supervised data available. It indicates that CLEVE is especially helpful for low-resource EE tasks, which is common since the expensive event annotation.
\subsection{Effect of AMR Parsers}
\label{sec:analysis_AMR}
\begin{table}[t!]
\small
\centering
\setlength{\tabcolsep}{3.5pt}
{
\begin{tabular}{l|c|cc|c}
\toprule
\multirow{2}{*}{} & \multicolumn{1}{c|}{\textbf{AMR 1.0}} & \multicolumn{2}{c|}{\textbf{ACE 2005}}                     & \multicolumn{1}{c}{\textbf{MAVEN}}                       \\ \cmidrule{2-5} 
                 & \multicolumn{1}{c|}{\textbf{Parsing}} & \multicolumn{1}{c}{\textbf{ED}} & \multicolumn{1}{c|}{\textbf{EAE}} & \multicolumn{1}{c}{\textbf{ED}}       \\ \midrule
\citet{wang-etal-2015-boosting}           & $62.0$   &        $79.8$      &          $61.1$       &          $68.5$         \\
\citet{xu-etal-2020-improving}       & $79.1$   &        $80.6$      &          $61.5$       &          $69.0$       \\ \bottomrule
\end{tabular}
}
\caption{Supervised results (F1,\%) on ACE 2005 and MAVEN of CLEVE using different AMR parsers, as well as the performance (F1,\%) of the parsers on AMR 1.0 (LDC2015E86) dataset.}
\label{tab:amr_ablation}
\end{table}
CLEVE relies on automatic AMR parsers to build self-supervision signals for large unsupervised data. Intuitively, the performance of AMR parsers will influence CLEVE performance. To analyze the effect of different AMR parsing performance, we compare supervised EE results of CLEVE models using the established CAMR~\cite{wang-EtAl:2016:SemEval} and a new state-of-the-art parser~\cite{xu-etal-2020-improving} during pre-training in Table~\ref{tab:amr_ablation}. We can see that a better AMR parser intuitively brings better EE performance, but the improvements are not so significant as the corresponding AMR performance improvement, which indicates that CLEVE is generally robust to the errors in AMR parsing.

\subsection{Effect of Pre-training Domain}
\label{sec:analysis_pre_domain}

\begin{table}[t!]
\small
\centering
\begin{tabular}{l|cc|c}
\toprule
\multirow{2}{*}{} & \multicolumn{2}{c|}{\textbf{ACE 2005}}                     & \multicolumn{1}{c}{\textbf{MAVEN}}                        \\ \cmidrule{2-4} 
                  & \multicolumn{1}{c}{\textbf{ED}} & \multicolumn{1}{c|}{\textbf{EAE}} & \multicolumn{1}{c}{\textbf{ED}}  \\ \midrule
NYT&      $\bm{79.8}$      &   $\bm{61.1}$          &     $68.5$      \\
\quad w/o semantic & $77.4$ & $59.7$ & $68.2$ \\
\quad w/o structure & $79.5$ & $60.7$ & $68.4$ \\
\midrule
Wikipedia         &      $79.1$      &            $60.4$          &     $\bm{68.8}$      \\
\quad w/o semantic & $77.3$ & $59.5$ & $68.4$ \\
\quad w/o structure & $78.8$ & $60.0$ & $68.6$ \\\bottomrule
\end{tabular}
\caption{Supervised results (F1,\%) on ACE 2005 and MAVEN of CLEVE pre-trained on different corpora.}
\label{tab:pre_domain}
\end{table}

Pre-training on similar text domains may further improve performance on corresponding downstream tasks~\cite{dontstoppretraining2020,gu-etal-2020-train}. To analyze this effect, we evaluate the supervised EE performance of CLEVE pre-trained on NYT and English Wikipedia in Table~\ref{tab:pre_domain}. We can see pre-training on a similar domain (NYT for ACE 2005, Wikipedia for MAVEN) surely benefits CLEVE on corresponding datasets. On ACE 2005, although Wikipedia is $2.28$ times as large as NYT, CLEVE pre-trained on it underperforms CLEVE pre-trained on NYT (both in the news domain). Moreover, we can see the in-domain benefits mainly come from the event semantics rather than structures in CLEVE framework (from the comparisons between the \textbf{w/o semantic} and \textbf{w/o structure} results). It suggests that we can develop domain adaptation techniques focusing on semantics for CLEVE, and we leave it to future work.

\section{Conclusion and Future work}
In this paper, we propose CLEVE, a contrastive pre-training framework for event extraction to utilize the rich event knowledge lying in large unsupervised data.
Experiments on two real-world datasets show that CLEVE can achieve significant improvements in both supervised and unsupervised ``liberal'' settings. In the future, we will (1) explore other kinds of semantic structures like the frame semantics and (2) attempt to overcome the noise in unsupervised data brought by the semantic parsers.

\section*{Acknowledgement}
This work is supported by the National Natural Science Foundation of China Key Project (NSFC No. U1736204), grants from Beijing Academy of Artificial Intelligence (BAAI2019ZD0502) and the Institute for Guo Qiang, Tsinghua University (2019GQB0003). This work is also supported by the Pattern Recognition Center, WeChat AI, Tencent Inc. We thank Lifu Huang for his help on the unsupervised experiments and the anonymous reviewers for their insightful comments.

\section*{Ethical Considerations}
We discuss the ethical considerations and broader impact of the proposed CLEVE method in this section: (1) \textbf{Intellectual property}. NYT and ACE 2005 datasets are obtained from the linguistic data consortium (LDC), and are both licensed to be used for research. MAVEN is publicly shared under the CC BY-SA 4.0 license\footnote{\url{https://creativecommons.org/licenses/by-sa/4.0/}}. The Wikipedia corpus is obtained from the Wikimedia dump\footnote{\url{https://dumps.wikimedia.org/}}, which is shared under the CC BY-SA 3.0 license\footnote{\url{https://creativecommons.org/licenses/by-sa/3.0/}}. The invited expert is fairly paid according to agreed working hours. (2) \textbf{Intended use}. CLEVE improves event extraction in both supervised and unsupervised settings, i.e., better extract structural events from diverse raw text. The extracted events then help people to get information conveniently and can be used to build a wide range of application systems like information retrieval~\cite{glavavs2014event} and knowledge base population~\cite{ji-grishman-2011-knowledge}. As extracting events is fundamental to various applications, the failure cases and potential bias in EE methods also have a significant negative impact. We encourage the community to put more effort into analyzing and mitigating the bias in EE systems. Considering CLEVE does not model people's characteristics, we believe CLEVE will not bring significant additional bias. (3) \textbf{Misuse risk.} Although all the datasets used in this paper are public and licensed, there is a risk to use CLEVE methods on private data without authorization for interests. We encourage the regulators to make efforts to mitigate this risk. (4) \textbf{Energy and carbon costs.} To estimate the energy and carbon costs, we present the computing platform and running time of our experiments in Appendix~\ref{app:training_details} for reference. We will also release the pre-trained checkpoints to avoid the additional carbon costs of potential users. We encourage the users to try model compression techniques like distillation and quantization in deployment to reduce carbon costs.

\bibliographystyle{acl_natbib}
\bibliography{acl2021}

\appendix

\section{Downstream Adaptation of CLEVE}
\label{app:adaption}
In this section, we introduce how to adapt pre-trained CLEVE to make the event semantic and structure representations work together in downstream event extraction settings in detail, including supervised EE and unsupervised ``liberal'' EE. 

\subsection{Supervised EE}
In supervised EE, we fine-tune the pre-trained text encoder and graph encoder of CLEVE with annotated data. We formulate both event detection (ED) and event argument extraction (EAE) as multi-class classification tasks. An instance is defined as a sentence with a trigger candidate for ED, and a sentence with a given trigger and an argument candidate for EAE. The key question here is how to obtain features of an instance to be classified.

For the event semantic representation, we adopt dynamic multi-pooling to aggregate the embeddings produced by text encoder into a unified semantic representation $\mathbf{x}_{sem}$ following previous work~\cite{chen2015event,wang-etal-2019-adversarial-training,wang-etal-2019-hmeae}. Moreover, we also insert special markers to indicate candidates as in pre-training (Section~\ref{sec:semantic}).
For the event structure representation, we parse the sentence into an AMR graph and find the corresponding node $v$ of the trigger/argument candidate to be classified. Following~\citet{qiu2020GCC}, we encode $v$ and its one-hop neighbors with the graph encoder to get the desired structure representation $\mathbf{g}_{str}$. The initial node representation is also obtained with the text encoder as introduced in Section~\ref{sec:structure}. 

We concatenate $\mathbf{x}_{sem}$ and $\mathbf{g}_{str}$ as the instance embedding and adopt a multi-layer perceptron along with $\mathrm{softmax}$ to get the logits. Then we fine-tune CLEVE with cross-entropy loss.

\subsection{Unsupervised ``Liberal'' EE}
Unsupervised ``liberal'' EE requires to discover event instances and event schemata only from raw text. We follow the pipeline of~\citet{huang-etal-2016-liberal} to parse sentences into AMR graphs and identify trigger and argument candidates with the AMR structures. We also cluster the candidates to get event instances and schemata with the joint constraint clustering algorithm~\cite{huang-etal-2016-liberal}, which requires semantic representations of the trigger and argument candidates as well as the event structure representations. The details of this clustering algorithm is introduced in Appendix~\ref{app:cluster}. Here we straightforwardly use the corresponding text span representations (Section~\ref{sec:semantic}) as semantic representations and encode the whole AMR graphs with the graph encoder to get desired event structure representations.

\section{Joint Constraint Clustering Algorithm}
\label{app:cluster}
In the unsupervised ``liberal'' event extraction~\citep{huang-etal-2016-liberal}, the joint constraint clustering algorithm is introduced to get trigger and argument clusters given trigger and argument candidate representations. CLEVE focuses on learning event-specific representations and can use any clustering algorithm. To fairly compare with~\citet{huang-etal-2016-liberal}, we also use the joint constraint clustering algorithm in our unsupervised evaluation. Hence we briefly introduce this algorithm here.

\subsection{Preliminaries} 
The input of this algorithm contains a trigger candidate set $T$ and an argument candidate set $A$ as well as their semantic representations $E_g^T$ and $E_g^A$, respectively. There is also an event structure representation $E_R^t$ for each trigger $t$. We also previously set the ranges of the numbers of resulting trigger and argument clusters: the minimal and maximal number of trigger clusters $K_{T}^{\mathrm{min}}$, $K_{T}^{\mathrm{max}}$ as well as the minimal and maximal number of argument clusters $K_{A}^{\mathrm{min}}$, $K_{A}^{\mathrm{max}}$. The algorithm will output the optimal trigger clusters $C^T=\{C_1^T,...,C_{K_T}^T\}$ and argument clusters $C^A=\{C_1^A,...,C_{K_A}^A\}$.

\subsection{Similarity Functions}
The clustering algorithm requires to define trigger-trigger similarities and argument-argument similarities. \citet{huang-etal-2016-liberal} first defines the constraint function $f$:
\begin{equation}
    \begin{aligned}
    f(\mathcal{P}_{1}, \mathcal{P}_{2})= \mathrm{log}(1+\frac{|\mathcal{L}_1\cap\mathcal{L}_2|}{|\mathcal{L}_1\cup\mathcal{L}_2|}).
    \end{aligned}
\end{equation}

When $\mathcal{P}_{1}$ and $\mathcal{P}_{2}$ are two triggers, $\mathcal{L}_i$ has tuple elements  $(\mathcal{P}_i,r,\mathrm{id}(a))$, which means the argument $a$ has a relation $r$ to trigger $\mathcal{P}_{i}$. $\mathrm{id}(a)$ is the cluster ID for the argument $a$. When $\mathcal{P}_{i}$ is arguments, $\mathcal{L}_i$ changes to corresponding triggers and semantic relations accordingly.

Hence the similarity functions are defined as:
\begin{equation}
    \small
    \begin{aligned}
    \mathrm{sim}(t_1,t_2)&= \lambda~\mathrm{sim}_{\mathrm{cos}}(E_g^{t_1},E_g^{t_2}) + f(t_1,t_2) \\ &+ (1-\lambda) \frac{\sum_{r\in R_{t_1}\cap R_{t_2}}\mathrm{sim}_{\mathrm{cos}}(E_r^{t_1}, E_r^{t_2})}{|R_{t_1}\cap R_{t_2}|} 
    , \\
    \\
    \mathrm{sim}(a_1,a_2) &= \mathrm{sim}_{\mathrm{cos}}(E_g^{a_1},E_g^{a_2}) + f(a_1,a_2)
    \end{aligned}
\end{equation}

where $E_g^t$ and $E_g^a$ are trigger and argument semantic representations, respectively. $R_t$ is the AMR relation set in the parsed AMR graph of trigger $t$. $E_r^t$ denotes the event structure representation of the node that has a semantic relation $r$ to trigger $t$ in the event structure. $\lambda$ is a hyper-parameter. $\mathrm{sim}_{\mathrm{cos}}(\cdot,\cdot)$ is the cosine similarity.

\subsection{Objective}
~\citet{huang-etal-2016-liberal} also defines an objective function $O(\cdot,\cdot)$ to evaluate the quality of trigger clusters $C^T=\{C_1^T,...,C_{K_T}^T\}$ and argument clusters $C^A=\{C_1^A,...,C_{K_A}^A\}$. It is defined as follows:
\begin{equation}
    \small
    \begin{aligned}
    O(C^T,C^A) &= D_{\mathrm{inter}}(C^T)+D_{\mathrm{intra}}(C^T) \\ &+ D_{\mathrm{inter}}(C^A)+D_{\mathrm{intra}}(C^A), \\
    D_{\mathrm{inter}}(C^\mathcal{P}) &= \sum_{i\neq j=1}^{K_{\mathcal{P}}}\sum_{u\in C_i^\mathcal{P}, v\in C_j^\mathcal{P}} \mathrm{sim}(\mathcal{P}_u,\mathcal{P}_v),\\
    D_{\mathrm{intra}}(C^\mathcal{P}) &= \sum_{i=1}^{K_{\mathcal{P}}} \sum_{u,v \in C_i^\mathcal{P}} (1-\mathrm{sim}(\mathcal{P}_u,\mathcal{P}_v)),
    \end{aligned}
\end{equation}
where $D_{\mathrm{inter}}(\cdot)$ measures the agreement across clusters, and $D_{\mathrm{intra}}(\cdot)$ measures the disagreement within clusters. The clustering algorithm iteratively minimizes the objective function.

\subsection{Overall Pipeline}
This algorithm updates its clustering results iteratively. At first, it uses the Spectral Clustering algorithm~\cite{vontutorial} to get initial clustering results. Then for each iteration, it updates clustering results and the best objective value using previous clustering results. It selects the clusters with the minimum $O$ value as the final result. The overall pipeline is shown in Algorithm~\ref{algo1}.

\begin{algorithm}[!t]
        \small
        \caption{Joint Constraint Clustering Algorithm}
    \KwInput{Trigger candidate set $T$, Argument candidate set $A$, their semantic representations $E_g^T$ and $E_g^A$, structure representations $E_R^t$ for each trigger $t$, the minimal and maximal number of trigger clusters $K_{T}^{\mathrm{min}}$, $K_{T}^{\mathrm{max}}$ as well as the minimal and maximal number of argument clusters $K_{A}^{\mathrm{min}}$, $K_{A}^{\mathrm{max}}$;}
    \KwOutput{Optimal trigger clusters $C^T=\{C_1^T,...,C_{K_T}^T\}$ and argument clusters $C^A=\{C_1^A,...,C_{K_A}^A\}$;}
        \begin{itemize}
        \item $O_{\mathrm{min}} = \infty, C^T = \emptyset, C^A = \emptyset$
        \item For $K_T = K_T^\mathrm{min}$ to $K_T^\mathrm{max}$, $K_A = K_A^\mathrm{min}$ to $K_A^\mathrm{max}$
        \begin{itemize}
            \item Clustering with Spectral Clustering Algorithm:
            \item $C_\mathrm{curr}^T = \mathrm{spectral}(T,E_g^T, E_R^T, K_T, C_\mathrm{curr}^A)$
            \item $C_\mathrm{curr}^A = \mathrm{spectral}(A,E_g^A, K_A)$
            \item $O_\mathrm{curr} = O(C_\mathrm{curr}^T,C_\mathrm{curr}^A)$
            \item if $O_\mathrm{curr} \leq O_\mathrm{min}$
            \begin{itemize}
                \item $O_{\mathrm{min}} = O_\mathrm{curr}, C^T = C_\mathrm{curr}^T, C^A = C_\mathrm{curr}^A$
            \end{itemize}
            \item while iterate time $\leq$ 10
            \begin{itemize}
                \item $C_\mathrm{curr}^T = \mathrm{spectral}(T,E_g^T, E_R^T, K_T, C_\mathrm{curr}^A)$
                \item $C_\mathrm{curr}^A = \mathrm{spectral}(A,E_g^A, K_A, C_\mathrm{curr}^T)$
                \item $O_\mathrm{curr} = O(C_\mathrm{curr}^T,C_\mathrm{curr}^A)$
                \item if $O_\mathrm{curr} \leq O_\mathrm{min}$
                \begin{itemize}
                    \item $O_{\mathrm{min}} = O_\mathrm{curr}, C^T = C_\mathrm{curr}^T, C^A = C_\mathrm{curr}^A$
                \end{itemize}
            \end{itemize}
        \end{itemize}
            
        \item return $O_{\mathrm{min}}, C^T, C^A$
        \end{itemize}
    \label{algo1}
\end{algorithm}

\section{Subgraph Sampling}
\label{app:subgraph_sampling}
In the AMR subgraph discrimination task of event structure pre-training, we need to sample subgraphs from the parsed AMR graphs for contrastive pre-training. Here we adopt the subgraph sampling strategy introduced by~\citet{qiu2020GCC}, which consists of the random walk with restart (RWR), subgraph induction and anonymization:

\begin{itemize}
    \item \textbf{Random walk with restart} first randomly chooses a starting node (the \textit{ego}) from the AMR graph to be sampled from. The \textit{ego} must be a root node, i.e., there is no directed edge in the AMR graph pointing to the node. Then we treat the AMR graph as an undirected graph and do random walks starting from the \textit{ego}. At each step, the random walk with a probability to return to the \textit{ego} and restart. When all the neighbouring nodes of the current node have been visited, the RWR ends.
    
    \item \textbf{Subgraph induction} is to take the induced subgraph of the node set obtained with RWR as the sampled subgraphs.
    
    \item \textbf{Anonymization} is to randomly shuffle the indices of the nodes in the sampled subgraph to avoid overfitting to the node representations.
\end{itemize}

In our event structure pre-training, we take subgraphs of the same sentence (AMR graph) as positive pairs. But, ideally, the two subgraphs in a positive pair should be taken from the same event rather than only the same sentence. However, it is hard to unsupervisedly determine which parts of an AMR graph belong to the same event. We think this task is almost as hard as event extraction itself. The rule used in the event semantic pre-training only handles the \texttt{ARG}, \texttt{time} and \texttt{location} relations, and for the other about $100$ AMR relations, we cannot find an effective method to determine which event their edges belong to. Hence, to take advantage of all the structure information, we adopt a simple assumption that the subgraphs from the same sentence express the same event (or at least close events) to design the subgraph sampling part here. We will explore more sophisticated subgraph-sampling strategies in our future work.

\begin{table}[t!]
\small
\centering
\setlength{\tabcolsep}{3.2pt}
{
\begin{tabular}{l|c}
\toprule
Batch size      & $40$    \\
Learning rate   & $1\times 10^{-5}$   \\
Adam $\epsilon$ & $1\times 10^{-8}$    \\
Adam $\beta_1$  & $0.9$     \\
Adam $\beta_2$  & $0.999$    \\
Trigger negative sampling size $m_t$ & $9$ \\
Argument negative sampling size $m_a$ & $30$ \\
Max sequence length & $128$ \\
\#parameters of text encoder & $355$M \\
\bottomrule
\end{tabular}}
\caption{Hyperparameters for the event semantic pre-training.}
\label{table:hproberta}
\end{table}

\begin{table}[t!]
\centering
\setlength{\tabcolsep}{3.5pt}
{
\begin{tabular}{l|c}
\toprule
Batch size      & $1024$    \\
Restart probability    & $0.8$     \\
Temperature     & $0.07$    \\
Warmup steps    & $7,500$   \\
Weight decay    & $1\times 10^{-5}$    \\
Training steps  & $75,000$  \\
Learning rate   & $0.005$   \\
Adam $\epsilon$ & $1\times 10^{-8}$    \\
Adam $\beta_1$  & $0.9$     \\
Adam $\beta_2$  & $0.999$    \\
Number of layers & $5$      \\
Dropout rate    &  $0.5$    \\
Hidden dimensions    &  $64$     \\
\#parameters of graph encoder & $0.2$M \\
\bottomrule
\end{tabular}}
\caption{Hyperparameters for the event structure pre-training.}
\label{table:hpgcc}
\end{table}

\section{Hyperparameter Setup}
\label{app:hyperparameter}
\subsection{Pre-training Hyperparameters}

During pre-training, we manually tune the hyperparameters and select the models by the losses on a held-out validation set with $1,000$ sentences. The event structure pre-training hyperparameters mainly follow the E2E model of~\citet{qiu2020GCC}. Table~\ref{table:hproberta} and Table~\ref{table:hpgcc} show the best-performing hyper-parameters used in experiments of the event semantic pre-training and event structure pre-training, respectively.

\begin{table}[t!]
\centering
\setlength{\tabcolsep}{3.5pt}
{
\begin{tabular}{l|c}
\toprule
Batch size      & $40$    \\
Training epoch  & $30$    \\
Learning rate  &  $1\times 10^{-5}$    \\
Adam $\epsilon$ & $1\times 10^{-8}$    \\
Adam $\beta_1$  & $0.9$     \\
Adam $\beta_2$  & $0.999$    \\
Max sequence length & $128$ \\
\bottomrule
\end{tabular}}
\caption{Fine-tuning hyperparameters for CLEVE and RoBERTa in the supervised setting.}
\label{table:hpfinetune}
\end{table}

\subsection{Fine-tuning Hyperparameters}

CLEVE in the unsupervised ``liberal'' setting directly uses the pre-trained representations and hence does not have additional hyperparameters. For the fine-tuning in the supervised setting, we manually tune the hyperparameters by $10$ trials. In each trial, we train the models for $30$ epochs and select models by their F1 scores on the validation set. Table~\ref{table:hpfinetune} shows the best fine-tuning hyperparameters for CLEVE models and RoBERTa. For the other baselines, we take their reported results.

\section{Training Details}
\label{app:training_details}
For reproducibility and estimating energy and carbon costs, we report the computing infrastructures and average runtime of experiments as well as validation performance.

\subsection{Pre-training Details}
For pre-training, we use 8 RTX 2080 Ti cards. The event semantic pre-training takes $12.3$ hours. The event structure pre-training takes $60.2$ hours.

\subsection{Fine-tuning/Inference Details}
During the fine-tuning in the supervised setting and the inference in the unsupervised ``liberal'' setting, we also use 8 RTX 2080 Ti cards.

For the supervised EE experiments, Table~\ref{tab:supervised_acevalidation} and Table~\ref{tab:supervised_MAVENval} show the runtime and the results on the validation set of the model implemented by us. 

In the unsupervised "liberal" setting, we only do inference and do not involve the validation. We report the runtime of our models in Table~\ref{tab:unsupervised_runtime}.

\begin{table}[!t]
  \centering
   \small
   \setlength{\tabcolsep}{2.4pt}
   {
   \scalebox{0.9}{
    \begin{tabular}{l|ccc|ccc|c}
    \toprule
          & \multicolumn{3}{c|}{\textbf{ED}} & \multicolumn{3}{c|}{\textbf{EAE}} & \multicolumn{1}{c}{\textbf{Runtime}} \\
    \midrule
    \textbf{Metric} & \textbf{P} & \textbf{R} & \textbf{F1}    & \textbf{P} & \textbf{R} & \textbf{F1} & \textbf{mins}\\
    \midrule
    RoBERTa & $72.9$ & $75.2$ & $74.0$ & $54.3$ & $62.6$ & $58.2$ & $344$\\
    \midrule
    CLEVE & $73.7$ & $79.4$ & $76.4$ & $56.2$ & $66.0$ & $60.7$ & $410$\\
    \quad w/o semantic & $72.1$ & $77.9$ & $74.9$ & $54.5$ & $65.6$ & $59.5$ & $422$\\
    \quad w/o structure & $73.2$ & $80.2$ & $76.5$ & $56.3$ & $65.4$ & $60.5$ & $355$\\
    \quad on ACE (golden) & $71.0$ & $77.1$ & $73.9$ & $55.0$ & $65.8$ & $59.9$ & $401$ \\
    \quad on ACE (AMR) & $70.2$ & $77.3$ & $73.6$ & $54.1$ & $65.5$ & $59.3$ & $408$\\
    \bottomrule
    \end{tabular}
    }
    }
    \caption{Supervised EE performance (\%) of various models on ACE 2005 validation set and the models' average fine-tuning runtime.}
  \label{tab:supervised_acevalidation}%
\end{table}

\begin{table}[t!]
\small
  \centering
    \begin{tabular}{l|ccc|c}
    \toprule
          & \multicolumn{3}{c|}{\textbf{ED}} & \multicolumn{1}{c}{\textbf{Runtime}}\\
    \midrule
    \textbf{Metric} & \textbf{P} & \textbf{R} & \textbf{F1} & \textbf{mins}\\
    \midrule
    RoBERTa & $65.3$ & $71.4$ & $68.2$ & $530$ \\
    \midrule
    CLEVE & $66.1$ & $70.2$ & $68.1$ & $572$  \\
    \quad w/o semantic & $66.5$ & $69.3$ & $67.9$ & $588$ \\
    \quad w/o structure & $65.4$ & $71.7$ & $68.4$ & $549$\\
    \bottomrule
    \end{tabular}%
    \caption{Supervised EE performance (\%) of various models on MAVEN validation set and the models' average fine-tuning runtime.}
  \label{tab:supervised_MAVENval}%
\end{table}

\begin{table}[t!]
 \centering
\begin{tabular}{l|c|c}
\toprule
\multirow{2}{*}{} & \multicolumn{1}{c|}{\textbf{ACE 2005}}                     & \multicolumn{1}{c}{\textbf{MAVEN}}     \\ 
\midrule
RoBERTa            & $12$~mins   &        $29$~mins     \\
RoBERTa+VGAE       & $17$~mins   &        $36$~mins     \\ 
\midrule
CLEVE               & $15$~mins & $33$~mins   \\
\quad w/o semantic  & $15$~mins & $32$~mins\\
\quad w/o structure & $14$~mins & $26$~mins \\
\bottomrule
\end{tabular}
    \caption{Average runtime of various models on ACE and MAVEN for the unsupervised ``liberal'' EE.}
    \label{tab:unsupervised_runtime}
\end{table}

\quad

\quad

\quad

\quad

\quad

\quad

\quad

\quad

\end{document}